\documentclass[conference,hidelinks]{IEEEtran}
\IEEEoverridecommandlockouts

\usepackage{cite}
\usepackage{amsmath,amssymb,amsfonts}
\usepackage{algorithmic}
\usepackage{graphicx}
\usepackage{textcomp}
\usepackage{xcolor}
\usepackage{booktabs}
\usepackage{tabularx}
\usepackage{multirow}
\usepackage{tikz}
\usepackage{orcidlink}
\usetikzlibrary{arrows.meta,positioning,fit,calc}
\def\BibTeX{{\rm B\kern-.05em{\sc i\kern-.025em b}\kern-.08em
    T\kern-.1667em\lower.7ex\hbox{E}\kern-.125emX}}
\newcolumntype{Y}{>{\raggedright\arraybackslash}X}

\begin{document}
\bstctlcite{IEEEexample:BSTcontrol}
\title{Checking the Checkers: A Synthesis of Verifier Noise and Robustness in RLVR}

\author{
    \IEEEauthorblockN{Savyasachi Gupta\,\orcidlink{0000-0002-9080-2028}\IEEEauthorrefmark{1}\thanks{\IEEEauthorrefmark{1}Both authors contributed equally to this work.}}
    \IEEEauthorblockA{\textit{Independent Researcher} \\
    TX, USA \\
    gsavya10@gmail.com}
    \and  
    \IEEEauthorblockN{Aniruddha Srinivas Joshi\,\orcidlink{0000-0002-1989-7597}\IEEEauthorrefmark{1}}
    \IEEEauthorblockA{\textit{Independent Researcher} \\
    CA, USA \\
    aniruddha980@gmail.com}
}

\maketitle

\begin{abstract}
Reinforcement learning with verifiable rewards (RLVR) has become a principal recipe for improving the reasoning of large language models (LLMs). It rests on the assumption that the checker scoring each response is itself correct. In practice, rule-based checkers reject correct answers in unexpected formats, unit tests accept incorrect programs, and LLM judges can be led to approve empty or adversarial responses. This work surveys 58 studies that measure, analyze, or mitigate such verifier errors. We ask what errors verifiers make, how they affect training, which mitigations work against which errors, and how the findings extend to self-generated, rubric- and judge-based, and verifier-free rewards. The evidence suggests that the structure of an error matters more than its rate. Random errors mostly slow learning, though under a fixed budget or KL regularization they also lower final accuracy. False positives that the policy can learn to trigger can cause plateaus or collapse, sometimes even with a low initial error rate. Whether a verifier wrongly rejects or accepts depends on what it checks and who built it, not on its family. Its errors shift as the policy improves, and most mitigations have been tested against only one kind of error. We compile the findings into comparison tables, reconcile apparent conflicts in the literature, and propose a reporting protocol to make future results comparable.
\end{abstract}


\section{Introduction}
Reinforcement learning (RL) post-training is now central to building large language models (LLMs) that reason well. Reinforcement learning with verifiable rewards (RLVR), a term introduced with the T\"ulu~3 post-training recipe~\cite{lambert2025tulu3}, replaces the learned reward model of classical RL from human feedback with a programmatic check: the final answer of a response is extracted and compared against a reference, or the generated program is executed against a set of unit tests. The approach was popularized by DeepSeek-R1~\cite{guo2025deepseekr1}, which paired such rewards with Group Relative Policy Optimization (GRPO)~\cite{shao2024deepseekmath}, and it now underpins most open reasoning models. Its appeal is easy to see. The reward is cheap to compute, it needs no human annotation, and it has been widely assumed to be difficult to game~\cite{guo2025deepseekr1,cui2025prime,srivastava2026rlsurvey}.

However, a verifier is not an oracle. It is a program, and sometimes another model, with failure modes of its own. The answers ``1/2'' and ``0.5'' are equal, yet a string-matching checker may disagree. A recent audit found that four configurations of widely used math verifiers accept anywhere between 53.8\,\% and 95.2\,\% of rewrites of the reference answer that are mathematically identical to it~\cite{xin2026category}. In code, an iterative attacker found a known-incorrect patch that passes on 28.5\,\% of sampled SWE-bench Verified tasks~\cite{rajan2026hackability}, and LLM judges can be made to approve a response that consists of nothing but a colon~\cite{zhao2025onetoken}. Since RL optimizes for what the verifier rewards rather than what its designer intended, verifier errors become part of the policy behavior. The problem grows as RLVR is extended from mathematics and code into medicine, science, and open-ended writing, where verification relies on rubrics and LLM judges whose errors include misspecified criteria as well as wrong verdicts~\cite{mahmoud2026rubrichacking,pan2026rubriceval}.

The literature on this problem has grown rapidly, yet it is scattered across communities that rarely cite one another: theoretical analyses of noisy policy gradients, audits of verification software, empirical studies of reward hacking, and work on label-free training, in which the model grades itself. It also appears contradictory, as some studies conclude that an imperfect verifier is good enough~\cite{plesner2026imperfect,lv2025climb}, while others find that noisy data is destructive~\cite{zhu2026noisy} or that certain errors cause complete collapse~\cite{egashira2026delay}; as we show, most of these results reconcile once the type of error is taken into account. 

Three surveys are closest to ours. Srivastava and Aggarwal~\cite{srivastava2026rlsurvey} review RL algorithms for LLMs and present RLVR as effectively eliminating reward hacking, since a policy cannot trick a compiler or a proof checker. Wang et al.~\cite{wang2026rewardhacking} explain reward hacking across RLHF, RLAIF and RLVR through proxy compression, and locate the weakness of RLVR in rewarding answers without checking the reasoning behind them. Gu et al.~\cite{gu2026judgesurvey} catalog the biases of LLM judges and attacks on them, including one-token attacks~\cite{zhao2025onetoken}, as threats to evaluation rather than as rewards a policy learns to trigger. None of them measures how often verifiers err or in which direction, or how errors that the policy does not exploit affect training. We adopt the distinction between random and systematic errors from a single controlled study~\cite{egashira2026delay}, apply it across 58 studies, including natural verifiers and label-free rewards, and connect it to the $J$-scaling results~\cite{rad2026rate,cai2026noisy,elmansouri2026noisegrpo}. This yields findings that no single study reports: what decides the direction of a verifier's errors (F2), which errors each mitigation has been tested against (Section~\ref{sec:rq3}), and a reporting protocol (Table~\ref{tab:protocol}).

\begin{figure*}[t]
\centering
\resizebox{\textwidth}{!}{
\begin{tikzpicture}[
    font=\scriptsize,
    root/.style={draw, very thick, rounded corners=2pt, fill=gray!15, align=center, inner sep=3pt, font=\footnotesize\bfseries},
    branch/.style={draw, thick, rounded corners=2pt, fill=gray!8, align=center, text width=3.95cm, inner sep=2.5pt, minimum height=8mm},
    leaf/.style={draw, rounded corners=1.5pt, align=left, text width=3.95cm, inner sep=2pt, font=\scriptsize},
    edge/.style={thin}
  ]
  \node[root] (root) {Verifier noise in reinforcement learning with verifiable rewards};

  \node[branch, below=5mm of root, xshift=-6.225cm] (b1) {\textbf{RQ1: What errors verifiers make} (Sec.~IV)};
  \node[branch, right=2mm of b1] (b2) {\textbf{RQ2: How errors affect training} (Sec.~V)};
  \node[branch, right=2mm of b2] (b3) {\textbf{RQ3: Mitigating verifier\\errors} (Sec.~VI)};
  \node[branch, right=2mm of b3] (b4) {\textbf{RQ4: Rewards without a trusted reference} (Sec.~VII)};
  \foreach \b in {b1,b2,b3,b4} \draw[edge] (root.south) -- ++(0,-2mm) -| (\b.north);

  \node[leaf, below=2mm of b1] (l11) {\textbf{Answer checkers and references:} mostly false negatives from format and configuration; wrong or incomplete labels \cite{huang2026accuracy,xu2026tinyv,xin2026category,zhu2026noisy}};
  \node[leaf, below=1.2mm of l11] (l12) {\textbf{Execution and formal checkers:} weak tests, proxy checks and loopholes accept wrong work \cite{zhang2026leaky,rajan2026hackability,helff2026gaming,ammanamanchi2026faults}};
  \node[leaf, below=1.2mm of l12] (l13) {\textbf{Model-based verifiers:} lean towards accepting; exploitable false positives \cite{cai2026noisy,zhao2025onetoken,li2026advjudge,tiwari2026prm}};
  \node[leaf, below=1.2mm of l13] (l14) {\textbf{From audits to training:} errors correlate and shift as the policy improves \cite{xin2026independent,kohli2026ninejudges,zhou2026variation,cui2025prime}};

  \node[leaf, below=2mm of b2] (l21) {\textbf{Random errors:} sign of $J$ sets direction; policy, granularity and budget set cost \cite{rad2026rate,lv2025climb,plesner2026imperfect,elmansouri2026noisegrpo}};
  \node[leaf, below=1.2mm of l21] (l22) {\textbf{Exploitable false positives:} plateau, collapse or rebound \cite{egashira2026delay,li2026advjudge,huang2026accuracy,wu2026remedy,khalifa2026countdown}};
  \node[leaf, below=1.2mm of l22] (l23) {\textbf{Errors the policy cannot exploit:} bounded or gradual loss; false negatives slow learning \cite{zhang2026leaky,zhu2026noisy,yan2026verifybench,xu2026tinyv}};
  \node[leaf, below=1.2mm of l23] (l24) {\textbf{Spurious-reward debate:} clipping, contamination or model strength \cite{shao2026spurious,wu2026contamination,chen2026exploration}};

  \node[leaf, below=2mm of b3] (l31) {\textbf{Correcting the gradient:} noise-aware estimators for class-conditional errors \cite{cai2026noisy,elmansouri2026noisegrpo}};
  \node[leaf, below=1.2mm of l31] (l32) {\textbf{Stronger verifiers and structural checks:} backups, hardened judges, gating \cite{xu2026tinyv,chen2025xverify,li2026advjudge,helff2026gaming,agrawal2026verigate}};
  \node[leaf, below=1.2mm of l32] (l33) {\textbf{Data, environments and oracles:} audits, re-verification, hardened graders \cite{zhu2026noisy,zhang2026leaky,thaman2026rhb,egashira2026delay}};
  \node[leaf, below=1.2mm of l33] (l34) {\textbf{Detecting and suppressing hacking:} detectors, representation-level control, aggregation \cite{deshpande2026trace,wu2026remedy,zhao2026care}};

  \node[leaf, below=2mm of b4] (l41) {\textbf{Self-generated rewards:} sharpen and collapse at scale; hidden per-problem loss \cite{zuo2025ttrl,he2026unsupervised,lin2026extinction,ye2026omgrpo}};
  \node[leaf, below=1.2mm of l41] (l42) {\textbf{Sparse external verification:} proofs, tools or targeted labels \cite{chen2026juryrl,liao2026t3rl,wang2026selfbelief}};
  \node[leaf, below=1.2mm of l42] (l43) {\textbf{Rubric and judge rewards:} judge error and rubric misspecification \cite{mahmoud2026rubrichacking,gunjal2026rubrics,wang2026cherrl,liu2026reasoningjudges}};
  \node[leaf, below=1.2mm of l43] (l44) {\textbf{Verifier-free rewards:} probability of the reference answer \cite{zhou2026verifree,yu2026rlpr,huang2026darl}};

  \foreach \b/\l in {b1/l11,b2/l21,b3/l31,b4/l41} \draw[edge] (\b.south) -- (\l.north);
\end{tikzpicture}}
\caption{Taxonomy of the surveyed literature on verifier noise in RLVR, organized by research question. Each leaf lists representative studies.}
\label{fig:taxonomy}
\end{figure*}
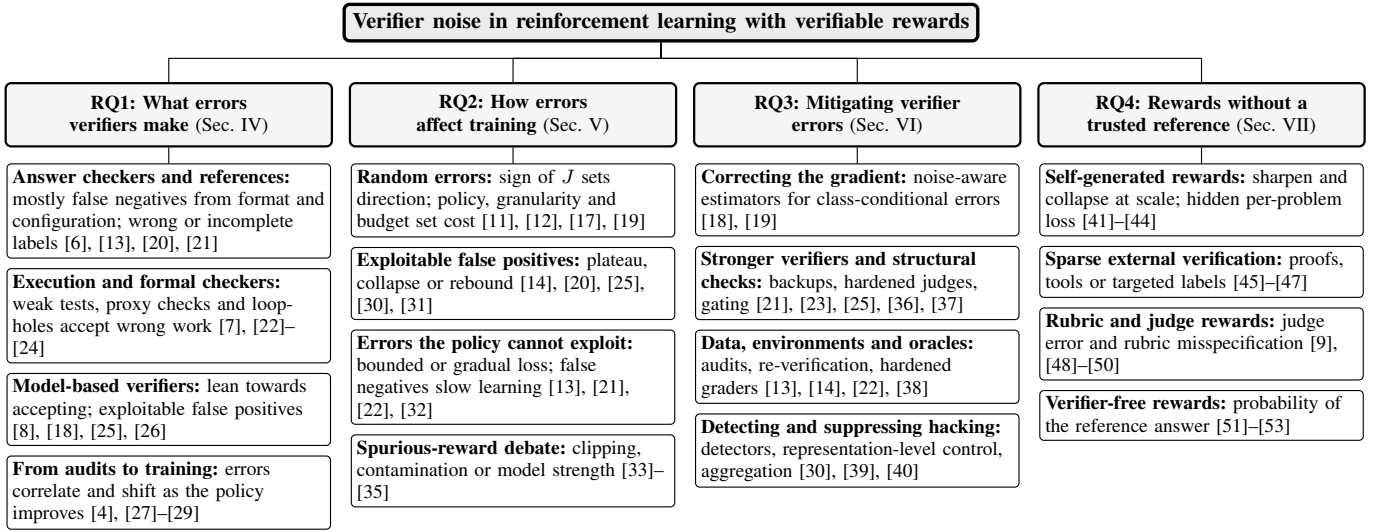

The key contributions of this paper are as follows:
\begin{itemize}
\item A structured review of 58 studies within the scope defined in Section~\ref{sec:method}, from each of which we extracted the verifier, the origin, direction, and structure of its errors, the training setup, the reported effect on training, and any mitigation with its cost.
\item A taxonomy (Fig.~\ref{fig:taxonomy}) that organizes the field into verifier errors, training responses, mitigation strategies, and rewards without a trusted reference.
\item A cross-study synthesis showing that whether the policy can exploit an error, more than how often the error occurs, decides the training outcome.
\item A reporting protocol and a list of open problems intended to make future studies comparable.
\end{itemize}

\section{Background}\label{sec:background}
\subsection{Reinforcement Learning with Verifiable Rewards}
In RLVR~\cite{lambert2025tulu3,guo2025deepseekr1,shao2024deepseekmath}, a policy $\pi_\theta$ samples $G$ responses $y_1,\dots,y_G$ to a prompt $x$, a verifier $V$ checks each against a reference $y^*$ (an answer key or a test suite) and gives it a reward $r_i = V(x, y_i, y^*) \in \{0,1\}$, and GRPO compares each response with its group as in Eq.~\ref{eq:adv}.
\begin{equation}
A_i = \frac{r_i - \mathrm{mean}(r_1,\dots,r_G)}{\mathrm{std}(r_1,\dots,r_G) + \epsilon}
\label{eq:adv}
\end{equation}
Dr.~GRPO drops the standard deviation and length normalization~\cite{liu2025drgrpo}, and DAPO filters out groups in which all responses share one reward~\cite{yu2025dapo}. Such groups give no gradient and are common even with a correct verifier~\cite{nie2026starvation}, and the group-relative advantage systematically underestimates the advantage of hard prompts and overestimates that of easy ones~\cite{yang2026biased}. Truncation handling also shapes the reward: penalizing truncated responses acts as reward noise~\cite{yu2025dapo}, and masking them helps at short generation budgets but not at long ones~\cite{liu2026acereason}. An idealized analysis shows that GRPO increases the probability of success above that of the reference policy~\cite{mroueh2025grpo}. As the analysis holds for any binary reward, we note that GRPO amplifies whatever the verifier accepts, including false positives.

\subsection{Verifiers}
We group verifiers by what they compare a response with. \emph{Answer checkers} extract the final answer, normalize it, and test it for equivalence with a reference, using string rules, numeric tolerances, or symbolic comparison~\cite{xin2026category,huang2026accuracy}. \emph{Execution-based checkers} run the response, as when a program is tested against unit tests~\cite{liu2023evalplus} or a proof is checked in Lean~\cite{ammanamanchi2026faults}. \emph{Model-based verifiers} are language models that judge a response against a reference, either prompted judges or trained verifiers~\cite{chen2025xverify,liu2025compassverifier}; process reward models apply the same idea to individual reasoning steps~\cite{lightman2024verify,cui2025prime}. All three depend on a trusted answer key or test and on a checker. A growing set of rewards drops one or the other, taking the signal from the policy's own majority vote~\cite{zuo2025ttrl}, from a judge scoring a rubric~\cite{gunjal2026rubrics}, or from the probability the policy assigns to the reference answer~\cite{zhou2026verifree}.

\subsection{A Model of Verifier Errors}\label{subsec:model_verifier_errors}
Let $r^*$ be the reward the designer intends, usually final-answer correctness. This is already a choice: if $r^*$ should reflect valid reasoning, a correct answer reached through flawed steps is a false positive~\cite{lightman2024verify}. Let $\tilde r$ be the reward the verifier emits. A false positive (FP) rewards an incorrect response with rate $\rho^{+} = P(\tilde r = 1 \mid r^* = 0)$, and a false negative (FN) rejects a correct one with rate $\rho^{-} = P(\tilde r = 0 \mid r^* = 1)$. If errors depend only on $r^*$, then
\begin{equation}
\mathbb{E}[\tilde r \mid r^*] = \rho^{+} + (1 - \rho^{+} - \rho^{-})\, r^* .
\label{eq:channel}
\end{equation}
The group baseline removes $\rho^{+}$, so the signal is scaled by $J = 1 - \rho^{+} - \rho^{-}$, the verifier's Youden index $\mathrm{TPR} - \mathrm{FPR}$~\cite{rad2026rate}. The same factor appears in work on noise-corrected GRPO~\cite{cai2026noisy,elmansouri2026noisegrpo} and, as the determinant of the reward confusion matrix, in earlier RL~\cite{wang2020perturbed}. It is exact for mean-centered advantages, while the standard deviation in Eq.~\ref{eq:adv} adds a further factor~\cite{elmansouri2026noisegrpo}.

Following Egashira et al.~\cite{egashira2026delay}, an error is \emph{random} when it is independent of the query and response given $r^*$, and \emph{systematic} when it depends on a property of the response, such as its format or a keyword. A policy can learn to trigger or avoid a systematic error but not a random one. We call a systematic FP \emph{exploitable} when the policy can produce its trigger often enough to reinforce it. Earlier RL work showed that corrupted rewards need not be unbiased~\cite{everitt2017corrupted}, that a non-equivalent proxy can be hacked~\cite{skalse2022defining}, and that a learned proxy is eventually overoptimized~\cite{gao2023scaling}. What is new in RLVR is that the proxy is often treated as exact; PRIME, for instance, calls exact match unhackable~\cite{cui2025prime}.

\section{Survey Methodology}\label{sec:method}
\subsection{Research Questions and Scope}
This work addresses the following research questions: 
\begin{enumerate}
    \item \textbf{RQ1:} What errors do the verifiers used in RLVR make, where do these errors come from, and how often do they occur?
    \item \textbf{RQ2:} How do verifier errors affect RLVR training, and what determines whether their effect is benign or harmful?
    \item \textbf{RQ3:} Which methods make RLVR robust to verifier errors, what do they cost, and against which kinds of error have they been shown to work?
    \item \textbf{RQ4:} How do these findings carry over to rewards that lack a trusted reference answer or a checker, namely self-generated, rubric- and judge-based, and verifier-free rewards?
\end{enumerate}

We cover work that trains or evaluates an LLM with an automatically computed reward and measures, analyzes, mitigates or exploits errors in that reward. Learned preference models from RL with human feedback are outside our scope unless they shed light on verifier errors.

\subsection{Search and Selection}
We searched arXiv, Google Scholar, and OpenReview for studies published up to September 2026. Keyword searches covered RLVR training with GRPO, noise-robust training and verifier-error correction, spurious rewards, vulnerabilities of LLM judges and reward models, reward hacking of unit tests, verifier models, and label-free and verifier-free training. We also screened the reference lists of four seed studies~\cite{egashira2026delay,plesner2026imperfect,xin2026category,xin2026independent}. Results were screened on title and abstract, and retained candidates were read in full; those within the scope above were included. This yielded 58 studies, from each of which we recorded the fields listed in the Introduction. A further 23 works, including three related surveys~\cite{wang2026rewardhacking,srivastava2026rlsurvey,gu2026judgesurvey}, are cited as background only. At the time of writing, most included preprints had been accepted at peer-reviewed venues but were not yet published in their respective proceedings.

\section{What Errors Verifiers Make (RQ1)}\label{sec:rq1}
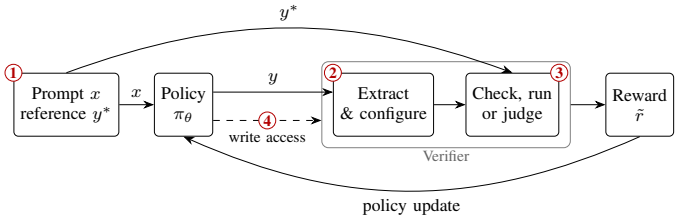
\begin{figure}[t]
\centering
\resizebox{\columnwidth}{!}{%
\begin{tikzpicture}[font=\footnotesize, >=Stealth, node distance=0.55cm,
  box/.style={draw, rounded corners=2pt, align=center, minimum height=1cm, inner sep=3pt},
  err/.style={circle, draw=red!70!black, fill=white, text=red!70!black, inner sep=0.6pt, font=\scriptsize\bfseries}]
\node[box] (data) {Prompt $x$\\ reference $y^*$};
\node[box, right=of data] (pol) {Policy\\ $\pi_\theta$};
\node[box, right=1.9cm of pol] (ext) {Extract\\ \& configure};
\node[box, right=0.5cm of ext] (chk) {Check, run\\ or judge};
\node[draw, gray, rounded corners=3pt, inner sep=5pt, fit=(ext)(chk),
      label={[font=\scriptsize, text=gray!80!black, inner sep=1pt]south:Verifier}] (ver) {};
\node[box, right=of ver] (rew) {Reward\\ $\tilde r$};
\draw[->] (data) -- node[above] {$x$} (pol);
\draw[->] ([yshift=2mm]pol.east) -- node[above] {$y$} ([yshift=2mm]ext.west);
\draw[->, dashed] ([yshift=-2.5mm]pol.east) -- node[below=1.5pt, font=\scriptsize] {write access} ([yshift=-2.5mm]pol.east -| ver.west);
\draw[->] (ext) -- (chk);
\draw[->] (ver) -- (rew);
\draw[->] (data.north) to[bend left=20] node[above] {$y^*$} (chk.north);
\draw[->] (rew.south) to[bend left=24] node[below] {policy update} (pol.south);
\node[err] at (data.north west) {1};
\node[err] at (ext.north west) {2};
\node[err] at (chk.north east) {3};
\node[err] at ($([yshift=-2.5mm]pol.east)!0.5!([yshift=-2.5mm]pol.east -| ver.west)$) {4};
\end{tikzpicture}}
\caption{Where errors enter the RLVR loop: (1) wrong or incomplete reference labels, (2) answer extraction and verifier configuration, (3) the checker, test suite, or judge itself, and (4) access that lets the policy alter the verifier or its environment.}
\label{fig:loop}
\end{figure}

Verifier errors enter the training loop at four points (Fig.~\ref{fig:loop}). Error direction depends more on what the checker actually tests and on who built it than on the family to which it belongs.

\subsection{Answer Checkers and Reference Labels}
Rule-based answer checkers fail mostly by rejecting correct answers. On mathematics, they reach a precision of at least 0.99, but their average recall is 0.86 to 0.93, and it falls to 0.35 to 0.58 on general-science questions~\cite{huang2026accuracy}. In an audit of 226K rejected responses, 38.5\,\% were in fact correct, most often because of differences in language, formatting, or notation~\cite{xu2026tinyv}. On MATH-500, a rule checker produced 10.4\,\% false negatives and no false positives~\cite{cai2026noisy}.

Many of these errors come from configuration. Four configurations accepted between 53.8\,\% and 95.2\,\% of the same certified-equivalent rewrites, two configurations of one library disagreed on 49.9\,\% of them, and whitespace and punctuation caused 93.0\,\% of one configuration's in-contract failures~\cite{xin2026category}. The same audit found a deterministic false positive: a cascade that accepts off-by-one answers whenever the reference is $10^4$ or larger. Answer extraction alone can shift MATH scores by up to 10 points~\cite{lambert2025tulu3}. Failures are also correlated within a group, with an intraclass correlation of 0.53 across eight rollouts, although shared prompt difficulty may explain part of this~\cite {xin2026independent}.

The reference can be wrong. When one study built noisy labels from a model's wrong answers, 16.4\,\% proved correct because gold answers were incomplete or equivalence checks failed, and 62\,\% of a Text2SQL sample had annotation errors~\cite{zhu2026noisy}. Wrong references affect 11\,\% of HumanEval solutions~\cite{liu2023evalplus} and about 20\,\% of PRM800K labels~\cite{shao2024deepseekmath}.

\subsection{Execution-Based Checkers}
Tests fail in the other direction. Expanding HumanEval's tests 80-fold lowers pass@k by up to 19.3--28.9\,\%~\cite{liu2023evalplus}, and the original MBPP tests accept 15.30\,\% of base-model samples that extended tests reject, though on inspection only about half of such false positives are wrong~\cite{zhang2026leaky}. In agentic code environments, an iterative attacker finds a known-incorrect patch that passes on 28.5\,\% of sampled SWE-bench Verified tasks (18.4\,\% in a single attempt) and at least 25.0\,\% of R2E-Gym tasks~\cite{rajan2026hackability}. Who writes the checker matters: LLM-written repair tests were defective 61.9\,\% of the time and rejected the reference patch, trading false positives for false negatives~\cite{rajan2026hackability}.

Deterministic checkers of a proxy property can be gamed. On an inductive-logic benchmark whose checker tests only the predicted labels, RLVR-trained models enumerate labels instead of inducing rules, on up to 74\,\% of hard tasks for one model~\cite{helff2026gaming}, and script-based graders for agentic tool use see exploit rates between 0\,\% and 22.3\,\% across 13 frontier models~\cite{thaman2026rhb}. Formal proofs are no exception, since the Lean kernel certifies only that a proof establishes the formal statement, not that the statement encodes the intended problem~\cite{ammanamanchi2026faults}. An audit of five benchmarks reports 4,833 findings (398 mechanically proven), from vacuous statements and harness loopholes that inflate scores to unprovable statements that deflate them~\cite{ammanamanchi2026faults}, and a pipeline that autoformalizes answers before proving them reaches 84.5\,\% precision~\cite{chen2026juryrl}.

\subsection{Model-Based Verifiers}
Model-based verifiers lean towards accepting. On MATH-500, an LLM verifier produced 33.6\,\% false positives and 16.2\,\% false negatives~\cite{cai2026noisy}, and samples on which verifiers disagreed were mostly incorrect, suggesting that judges over-accept~\cite{liu2025compassverifier}. Specialized verifiers are more accurate than general LLMs but miss valid variants~\cite{li2026verifybench}, and on hard items, general LLMs lose about 20 points of accuracy against 8 for the best systems~\cite{yan2026verifybench}. The lean is not universal: on calibration sets with synthetically corrupted negatives, AceMath-7B-RM has a false-positive rate of only 2.0\,\% but rejects 39.2\,\% of correct responses~\cite{elmansouri2026noisegrpo}.

These false positives are often exploitable. Single symbols or generic reasoning openers raise average false-positive rates to 80.6\,\% for Llama-3-70B, 66.8\,\% for Qwen2.5-72B, 14.6\,\% for GPT-4o, and 6.0\,\% for GPT-o1, with worst cases up to 95.1\,\%, and larger judges are not reliably safer~\cite{zhao2025onetoken}. Unlike earlier universal phrases found by an external attacker~\cite{raina2024judge}, tokens found by searching a judge's own output distribution push false-positive rates above 90\,\% on 22 of 24 judge and dataset pairs, though this is an upper bound: realistic appended tokens reach 27.5\,\% to 58\,\% on Llama-based judges and at most 23\,\% on Qwen- and Gemma-based ones~\cite{li2026advjudge}. Simple patterns fool generative verifiers, fine-tuned or not, in up to 77.9\,\% of cases, whereas discriminative verifiers stay robust~\cite{huang2026accuracy}. Process reward models reward surface form over logical validity: when a policy is trained against one with GRPO, 43\,\% of its reward gain vanishes once its responses are rephrased~\cite{tiwari2026prm}. Judge errors are also correlated, as nine frontier judges carry about two effective votes~\cite{kohli2026ninejudges}.

\subsection{From Static Audits to Training}
Almost every rate above is based on a fixed set of outputs, whereas during training, the verifier faces the policy's shifting outputs. The true negative rate of a reference-free verifier falls from 0.68 to 0.17 as the generator gets stronger, and its false negatives concentrate on hard problems (49.0\,\% of correct responses in the hardest quartile against 0.6\,\% in the easiest)~\cite{zhou2026variation}. An offline process reward model loses accuracy as RL proceeds~\cite{cui2025prime}. In theory, the accuracy reachable with an imperfect verifier is capped by its ROC curve near zero false positives, which low-compute measurements cannot predict~\cite{dorner2026rocnreroll}. Static audits may therefore understate how often a policy yields exploitable false positives for a verifier, although this has not been measured directly in RL.

\section{How Verifier Errors Affect Training (RQ2)}\label{sec:rq2}
Training outcomes depend less on how often a verifier errs than on what kind of error it makes. Random errors mostly slow learning; errors the policy can learn to trigger produce plateaus or collapse; and systematic errors tied to particular inputs, which the policy cannot trigger, gradually erode accuracy, or stay bounded (Table~\ref{tab:outcomes}).

\begin{table*}[t]
\caption{Training Outcomes Under Reference-Based Verifier Errors}
\label{tab:outcomes}
\centering
\scriptsize
\setlength{\tabcolsep}{3pt}
\begin{tabularx}{\textwidth}{@{}>{\raggedright\arraybackslash}p{1.75cm}>{\raggedright\arraybackslash}p{4.4cm}>{\raggedright\arraybackslash}p{5.3cm}Y@{}}
\toprule
\textbf{Study} & \textbf{Reward signal and error} & \textbf{Outcome} & \textbf{Key result} \\
\midrule
Egashira~\cite{egashira2026delay} & Rule checker; injected random FP/FN, systematic FN, exploitable FP & Random: delay (lower endpoint for OLMo at 50\,\% FNR); exploitable FP: plateau or collapse & Pass@1 0.6 with a ``python'' trigger vs.\ 85.6 clean \\
Rad~\cite{rad2026rate} & Unit tests; random flips across $J$ & $J>0$: delay or lower endpoint; $J=0$: no learning; $J<0$: collapse & At $J=0.3$: 16.0 FN-heavy vs.\ 14.6 FP-heavy (clean 20.8) \\
El Mansouri~\cite{elmansouri2026noisegrpo} & Rule and reward models; i.i.d.\ flips; no ratio clipping & Lower endpoint, strongly model-dependent & Losses of 0.2--30.7 points \\
Plesner~\cite{plesner2026imperfect} & Unit tests; symmetric flips; small model verifier & Group-level ${\le}15\,\%$: graceful; sample-level: lower endpoint; 4B verifier: plateau & Best reward 0.704 vs.\ 0.871 with a 30B verifier \\
Lv~\cite{lv2025climb} & Rule; question-wise flips; PPO & ${\le}40\,\%$: graceful or delay; 50\,\%: no learning; Llama-3.1-8B: collapse at 30\,\% & 72.02 at 40\,\% flips vs.\ 75.85 clean \\
Cai~\cite{cai2026noisy} & Rule; class-conditional and format-dependent noise & Uncorrected: lower endpoint and loss of general capability & ARC-C 73.9 $\rightarrow$ 43.9 (Llama-3.2-3B) \\
Xu~\cite{xu2026tinyv} & Rule; natural FN & Delay, then lower endpoint & 51.82 vs.\ 54.46 with FNs repaired \\
Zhu~\cite{zhu2026noisy} & Model-aligned wrong labels (the model's own wrong answers) & Lower endpoint, monotone in noise rate & $-9.0$ on MATH-500 at 100\,\% noise \\
Zhang~\cite{zhang2026leaky} & Leaky unit tests; natural FP & Proxy inflated; held-out accuracy unaffected & +8.37 training reward, 0.20 held-out \\
Huang~\cite{huang2026accuracy} & Rule vs.\ fine-tuned generative judge & Rule: lower endpoint; judge: divergence & Reward leaves oracle near step 450 \\
Agrawal~\cite{agrawal2026verigate} & PRM as outcome reward & Divergence, then collapse at 1.5B & GSM8K 71.11 $\rightarrow$ 31.16 \\
Zhao~\cite{zhao2025onetoken} & LLM judge; master-key FP & Collapse & 94.26\,\% of outputs become ``Thought Process:'' \\
Li~\cite{li2026advjudge} & Binary judge; FP discovered by the policy & Collapse in 5 of 10 seeds (MATH) & 88\,\% of false positives are empty outputs \\
Helff~\cite{helff2026gaming} & Extensional rule checker; exploitable FP & Plateau from about step 250 & Proxy--oracle gap of 3.5/10 at step 500 \\
Khalifa~\cite{khalifa2026countdown} & Test harness the policy can overwrite & SFT-primed: collapse (2 of 6 hack); RL only: plateau (2 of 8 hack) & Hacking peaks at 80--90\,\% of rollouts \\
Wu~\cite{wu2026remedy} & Evaluator the policy can overwrite & Rebound, then collapse & Hack rate 99.9\,\%, pass@1 1.2\,\% (Phi-4-mini) \\
\bottomrule
\multicolumn{4}{@{}p{\textwidth}@{}}{\scriptsize A plateau means the true reward stalls while the proxy rises; divergence means the true reward declines while the proxy rises; a rebound means hacking fails, retreats, and then succeeds.}
\end{tabularx}
\end{table*}

\subsection{Random Errors}
Under random errors, the sign of $J$ from Section~\ref{subsec:model_verifier_errors} sets the direction of learning. A mean-field analysis predicts learning at a rate scaled by $J$ when $J>0$, drift at $J=0$, and collapse when $J<0$. Experiments on code generation agree, although false positives cost more than false negatives at equal $J$~\cite{rad2026rate}. Policies can tolerate a great deal of such noise, even when it is applied per question rather than per response, thus correlated within a group. Qwen2.5-7B still climbs from 4.82\,\% to 72.02\,\% on MATH-500 when the rewards of 40\,\% of questions are flipped, against 75.85\,\% without noise~\cite{lv2025climb}, and group-level flips of up to 15\,\% keep the best validation reward within 2 points of the clean run~\cite{plesner2026imperfect}.

This tolerance has limits. It depends on the policy: Llama-3.1-8B collapses at 30\,\% flips~\cite{lv2025climb}, although $J = 0.4$ predicts only slower learning, and multimodal backbones differ widely in how much label corruption they withstand~\cite{xu2026twostage}. It depends on granularity: flipping just 10\,\% of individual samples, not whole groups, costs 3.0 to 4.7 points~\cite{plesner2026imperfect}. And it depends on the training setup. With fixed epochs, a KL penalty, and no ratio clipping, i.i.d.\ flips cost 0.2 to 30.7 points, but a short ablation restoring clipping raised the noisy baseline by about 6 points, so part of this cost may reflect the setup~\cite{elmansouri2026noisegrpo}.

\subsection{Exploitable False Positives}
Whether a systematic false positive is exploitable depends on whether the policy can reach its trigger. In controlled arithmetic tasks built with Reasoning Gym~\cite{stojanovski2025reasoninggym}, the drop in pass@1 follows the trigger's conditional advantage times its initial frequency~\cite{egashira2026delay}. A false positive for ``python'' drives pass@1 from 85.6\,\% to 0.6\,\%, one for the opener ``Certainly'' plateaus at 64.3\,\%, and one that fires only half the time is not exploited, as its conditional advantage is too low.

Natural verifiers behave likewise once the policy finds the trigger. With a judge susceptible to reasoning openers, 94.26\,\% of outputs became ``Thought Process:''~\cite{zhao2025onetoken}. Against a binary judge, 5 of 10 seeds collapsed on MATH, mostly onto empty outputs locked in within about 170 steps~\cite{li2026advjudge}. A fine-tuned generative verifier was hacked but a discriminative one was not~\cite{huang2026accuracy}, process reward models used as outcome rewards were hacked~\cite{agrawal2026verigate}, one of them despite resisting gradient attacks~\cite{tiwari2026prm}, and even a deterministic logic checker let proxy and true reward drift apart from about step 250~\cite{helff2026gaming}.

When the policy can modify the verifier, hacking can rebound: early attempts fail, the policy retreats, and hacking then succeeds, most readily when legitimate reward is scarce~\cite{wu2026remedy}. Hacking can also be inherited. After distillation on traces, only 1.2\,\% of which contained hacking, two of the six primed models (both Qwen) began hacking within 100 RL steps, and their true reward collapsed; the rest, including Llama-3.1-8B, hacked little or not at all. Without distillation, only two of eight models learned to hack, and their true reward plateaued~\cite{khalifa2026countdown}.

\subsection{Errors the Policy Cannot Exploit}
Systematic errors that the policy cannot turn into a reward behave much like noise. Leaky MBPP tests inflated training reward without changing held-out accuracy, and their false-positive mass did not grow over 800 steps~\cite{zhang2026leaky}. Model-aligned wrong labels, i.e., wrong reference answers drawn from the model's own errors, cost accuracy monotonically with the noise rate, from 1.5 points at 10\,\% noise to 9.8 at 100\,\% averaged over five math benchmarks, and push pass@$k$ below the base model for $k>1$~\cite{zhu2026noisy}. Judge accuracy sets the direction coarsely, as judges at or below 81\,\% accuracy pushed a policy below its starting point while judges at or above 92\,\% improved it~\cite{yan2026verifybench}, while uncorrected noise can erode capability outside the task~\cite{cai2026noisy}.

False negatives remove learning signal rather than reward wrong behavior, but at a cost: natural-rule false negatives slow early learning and lower the endpoint~\cite{xu2026tinyv}, and those spanning a domain produce a plateau~\cite{egashira2026delay}. Policies learn around them, answering in Chinese to escape a checker rejecting English~\cite{egashira2026delay} or appending {\small\texttt{\textbackslash boxed\{\}}} to pass Math-Verify~\cite{chen2025xverify}.

\subsection{The Spurious-Reward Debate}
Training Qwen2.5-Math-7B with random rewards improved MATH-500 accuracy by 21.4 points, nearly matching the 29.1 points gain from ground-truth rewards~\cite{shao2026spurious}. These gains did not transfer to other model families: Llama and OLMo2 improved little or degraded. They also largely vanished on AIME 2025. Three explanations compete: clipping that amplifies the model's prior for code-style reasoning~\cite{shao2026spurious}; contamination, since from 60\,\% of each problem Qwen2.5-Math-7B reconstructs 54.60\,\% of MATH-500 against 3.80\,\% for Llama3.1-8B and only correct rewards give stable gains on a clean benchmark~\cite{wu2026contamination}; and model strength, since the clipping bias is too small to carry a signal and gains appear on non-Qwen models too~\cite{chen2026exploration}. The studies differ in prompts, clipping, and models, so the question is open, but they agree that a strong prior can override the $J$ analysis and that benchmarks older than the model can make any reward look useful.

\subsection{What Decides the Outcome}
Three factors decide how training responds. The structure of the error sets the regime. The policy's prior decides whether an exploitable error is found, since a policy too weak to produce a trigger cannot exploit it~\cite{huang2026accuracy} and Llama-3.1-8B resisted hacking even after priming, unlike Qwen models of similar size~\cite{khalifa2026countdown}. The budget, regularization, and the supply of legitimate reward decide whether a delay becomes a loss~\cite{rad2026rate,elmansouri2026noisegrpo} and whether hacking rebounds~\cite{wu2026remedy}. 

\section{Mitigating Verifier Errors (RQ3)}\label{sec:rq3}
Mitigations act on the factors identified in Section~\ref{sec:rq2}: some correct the gradient for a known error rate, some replace or back up the verifier, some repair the data or the environment, and some detect or penalize hacking as it emerges. Almost all have been validated against a single error regime (Table~\ref{tab:mitigation}), and defenses against exploitable false positives have been tested inside RL in only seven studies~\cite{egashira2026delay,huang2026accuracy,agrawal2026verigate,li2026advjudge,helff2026gaming,khalifa2026countdown,wu2026remedy}, each on one task or a few models.

\begin{table}[t]
\caption{Mitigation Families and the Errors They Were Tested On}
\label{tab:mitigation}
\centering
\scriptsize
\setlength{\tabcolsep}{2.5pt}
\begin{tabularx}{\columnwidth}{@{}>{\raggedright\arraybackslash}p{2.35cm}>{\raggedright\arraybackslash}p{2.6cm}Y@{}}
\toprule
\textbf{Family} & \textbf{Tested against} & \textbf{Cost or gap} \\
\midrule
Noise-aware gradients~\cite{cai2026noisy,elmansouri2026noisegrpo} & Class-conditional noise; format FN; real reward models & Needs error rates; fails on model-aligned wrong labels~\cite{zhu2026noisy} \\
Noise-robust regularization~\cite{xu2026twostage} & Random label corruption & Backbone-dependent; extra hyperparameters \\
Stronger verifiers~\cite{xu2026tinyv,liu2025compassverifier,chen2025xverify,huang2026accuracy} & Natural rule FN; judge hacking & Extra inference; small RL gains; own FP rarely reported \\
Judge hardening~\cite{li2026advjudge,zhao2025onetoken} & Discovered and master-key FP & One policy--judge pair; offline only in~\cite{zhao2025onetoken} \\
Structural checks~\cite{helff2026gaming,agrawal2026verigate} & Enumeration shortcut; PRM hacking & Task-specific; outcome-verifier errors untouched \\
Data audits~\cite{zhu2026noisy,zhang2026leaky,khalifa2026countdown} & Wrong labels; leaky tests; inherited hacking & Covers only what is audited \\
Environment hardening~\cite{thaman2026rhb,rajan2026hackability,ammanamanchi2026faults} & Grader and harness exploits & Per-environment effort; mostly evaluation-time \\
Oracle access~\cite{egashira2026delay} & Plateau and collapse & Oracle on 10--50\,\% of steps; trigger persists \\
Detection, representation control~\cite{deshpande2026trace,wu2026remedy} & Hacking onset; test rewriting & Detects rather than prevents; white-box; one exploit \\
Judge aggregation~\cite{zhao2026care} & Correlated judges; master keys & Not tested in RL; small gains in~\cite{kohli2026ninejudges} \\
\bottomrule
\end{tabularx}
\end{table}

\subsection{Correcting the Gradient}
If the error rates are known, the gradient can be corrected. A backward correction needs both rates and becomes noisy as $J$ shrinks, whereas a forward correction needs only the false-negative rate, which can be estimated online by sending about 10\,\% of rule rejections to a small verifier; under format-dependent false negatives, it reaches 31.2 against 20.4 without correction~\cite{cai2026noisy}. With real verifiers, the gains are smaller, 1.2 to 2.8 points over a rule reward~\cite{cai2026noisy} and up to 6.7 points on math with reward models whose flip rates were estimated on a labeled calibration split~\cite{elmansouri2026noisegrpo}. Both corrections are applied after group normalization, which would otherwise cancel them, and both assume class-conditional noise. On wrong labels that follow the model's own errors, the forward correction failed even when given the true noise rates~\cite{zhu2026noisy}, and neither method has been tested against an exploitable false positive.

\subsection{Stronger Verifiers and Structural Checks}
The most common remedy is a better verifier, and offline accuracy buys less than one might expect. TinyV consults a 1.5B verifier only when the rule checker rejects, adding 2.6 to 2.8 points at 6\,\% extra step time, although its own false-positive rate is unreported~\cite{xu2026tinyv}. xVerify, far more accurate than Math-Verify as a classifier, improves on it by only 0.8 points as a reward~\cite{chen2025xverify}, and a hybrid rule-then-model verifier adds 2.3 points for 25.7\,\% more training time~\cite{huang2026accuracy}.

Hardening targets exploitable false positives directly. Fine-tuning a judge with LoRA on discovered tokens cut mean false positives during GRPO from 655 to 42 over ten seeds (median 168 to 40) and prevented all eight false-positive collapses across two datasets, yet tokens found by a fresh discovery run still reach false-positive rates of 8.75 to 31\,\% against the hardened judges~\cite{li2026advjudge}. Training on truncated responses labeled incorrectly brings master-key false positives to 0.1\,\%, but has been tested only offline~\cite{zhao2025onetoken}. Structural checks close specific loopholes: an isomorphic verifier keeps the proxy--oracle gap near zero in inductive logic~\cite{helff2026gaming}, and consulting a PRM only when every rollout fails avoids the hacking seen when the PRM is the reward~\cite{agrawal2026verigate}.

\subsection{Data, Environments and Oracles}
Several errors are cheaper to remove at their source than to correct in the gradient. Re-verifying labels was the only remedy for wrong references~\cite{zhu2026noisy}. A static audit of 16 samples per task predicts where leaky tests will reward wrong code, and repairing the 18 worst suites removes about half of that false-positive mass~\cite{zhang2026leaky}. Filtering distillation data on true rather than proxy reward removes inherited hacking~\cite{khalifa2026countdown}. Hardening agentic graders cut exploits from 6.5\,\% to 0.8\,\% with task success unchanged~\cite{thaman2026rhb}, and a Docker-gated loop repaired 9 of 11 hackable code tasks for \$3.60~\cite{rajan2026hackability}. Access to a trusted oracle helps but does not cure: alternating with an oracle every 10 steps turns plateaus into delays; preventing collapse requires the oracle every 2 steps; and even then, the policy keeps producing the trigger~\cite{egashira2026delay}.

\subsection{Detecting and Suppressing Hacking}
Detection tells practitioners when hacking has begun, but does not stop it. Comparing ten trajectories raises an LLM detector's macro-F1 from 0.45 to 0.63 on synthetic coding-agent hacks, but semantic hacks stay at 0.0 to 0.4~\cite{deshpande2026trace}. Probes on internal activations can flag hacking learned through fine-tuning~\cite{wilhelm2026monitoring}, and acting on the policy's own representations goes further. A shortcut direction extracted from contrastive sentence pairs tracks hacking during training, and discounting the advantage of rollouts that score high on it cuts Phi-4-mini's hack rate from 99.9\,\% to 24.9\,\% and lifts LeetCode pass@1 from 1.2\,\% to 12.0\,\%, whereas suppressing the same direction at generation time only lowers the hack rate to 72.3\,\%~\cite{wu2026remedy}. For correlated judges, confounder-aware aggregation beats majority voting on most benchmarks and neutralizes two master keys~\cite{zhao2026care}, but in another study, aggregation closed at most 11\,\% of the gap to independent voting~\cite{kohli2026ninejudges}, and neither has been tested inside RL.

No mitigation has been shown to handle random, systematic, and exploitable errors simultaneously, and each defense against exploitable false positives rests on a single policy, judge, or task. Section~\ref{sec:rq4} asks how far these remedies carry when the reference itself is no longer trusted.

\section{Rewards Without a Trusted Reference (RQ4)}\label{sec:rq4}
When the reference comes from the policy, a rubric, or a judge, or when no checker is used, the earlier findings carry over with one change: the reward can no longer be checked against an independent source, so errors that such a check would expose become systematic by construction.

\subsection{Self-Generated Rewards}
When the group's majority answer serves as the label~\cite{zuo2025ttrl}, label error and reward error come apart. At the start of training, only 37\,\% of majority labels were correct, yet 92\,\% of rewards were, because wrong answers rarely match a wrong majority~\cite{zuo2025ttrl}. The majority itself is wrong on 25.85\,\% of MATH-500 and 73.33\,\% of AIME 2024 questions in one study~\cite{liao2026t3rl}. The regime decides the outcome. On small or test-time sets, such training helps, lifting Qwen2.5-Math-7B from 12.9 to 40.2 on AIME 2024~\cite{zuo2025ttrl}. At the training-corpus scale, every intrinsic reward studied sharpens the initial distribution and eventually collapses, even in the most stable settings, by roughly 1,000 steps; in one majority-vote run, reward accuracy fell from about 0.80 to 0.48 within about 260 steps while the proxy reward kept rising~\cite{he2026unsupervised}. Majority-vote baselines collapsed in several studies~\cite{yu2025restrain,zhang2026corewarding,wang2026selfbelief}, in one case with rollout accuracy falling from 44.0\,\% to 2.4\,\% while training reward rose~\cite{ye2026omgrpo}.

Damage can also hide in aggregates. Pooled over three models, 21.6\,\% of problems were lost and only 0.7\,\% newly learned while pass@1 stayed flat or rose~\cite{lin2026extinction}, and about two-thirds of one audio gain came from a single phoneme template~\cite{vakada2026labelfree}. Coupling the reward to the online policy is no clear fix: a frozen self-certainty scorer was exploited near step 100 while the online one held for 150 steps~\cite{zhao2026intuitor}, but self-certainty collapsed in longer runs elsewhere~\cite{zhang2026corewarding,he2026unsupervised}.

Sparse external verification closes most of the gap. A Lean proof gate on the majority answer matched ground-truth reward on nine-benchmark averages for three models (e.g., 35.40 vs.\ 35.56)~\cite{chen2026juryrl}, labels bought for 20\,\% of problems prevented collapse when targeted but not when random~\cite{wang2026selfbelief}, and soft vote weights from a tool verifier with only 25 to 39\,\% precision beat majority-vote training on all 15 model and benchmark pairs, whereas hard filtering on the same verifier hurt~\cite{liao2026t3rl}. Stabilizers without any external signal delay collapse~\cite{zhang2026corewarding,yu2025restrain,ye2026omgrpo,du2026dualconsensus,wu2026spine} but cannot correct a consensus that is systematically wrong~\cite{wu2026spine}.

\subsection{Rubric and Judge Rewards}
Rubric judges, like answer judges, fail mostly by accepting. Against a panel of frontier models, a weak judge had 10.3\,\% false positives and 6.8\,\% false negatives on medical rubrics and 19.8\,\% and 4.4\,\% on science~\cite{mahmoud2026rubrichacking}, and on the harder half of disputed rubric items, balanced accuracy spans 55.65\,\% to 84.81\,\% across 18 judges~\cite{pan2026rubriceval}. The rubric adds an error of its own, since rubrics written without reference answers train worse policies (32.0 vs.\ 35.9 on HealthBench)~\cite{gunjal2026rubrics}. In training, the share of newly credited criteria that were wrong grew from 39\,\% to 65\,\% under a weak medical judge. A strong judge showed no such trend, yet rubric-free judges still preferred the base model on 78.4\,\% of prompts, so a better verifier does not fix a misspecified rubric~\cite{mahmoud2026rubrichacking}.

Exploitation again depends on discoverability. Injected judge biases were exploited after 68 to 478 steps, with less task-entangled biases exploited later, and two of the eight were never exploited~\cite{wang2026cherrl}. Non-reasoning judges were hacked at every size, while reasoning judges moved the hacking to the gold judge, which the policy's outputs also fooled, reaching an 89.6\,\% win rate on Arena-Hard-V2 creative writing~\cite{liu2026reasoningjudges}. Negated rubrics drive scores from 23.4 to 2.9, while incomplete ones degrade gracefully~\cite{zhou2026ruscarl}, a split that, in our reading, mirrors $J<0$ versus a smaller positive $J$. Decomposing the reward and auditing training runs both help: a checklist judge is less accurate than specialized reward models on RewardBench yet trains better policies~\cite{viswanathan2025checklists}, and a judge-blind auditing agent flagged all six hacked runs~\cite{wang2026cherrl}.

\subsection{Verifier-Free Rewards}
Verifier-free methods drop the checker and reward the policy's own probability of the reference answer. VeriFree does so after the policy's reasoning, which equals exact-match RLVR in expectation when there is a single correct answer~\cite{zhou2026verifree}. In our reading, it thereby trades the false positives of model verifiers for the false negatives of exact match, and adding equivalent answers helped only slightly~\cite{zhou2026verifree}. RLPR rewards the mean token probability instead and reaches a ROC-AUC of about 0.90 against human labels on general data, against 0.61 for a rule verifier and 0.69 for a model verifier~\cite{yu2026rlpr}. Probability rewards still under-credit correct paraphrases~\cite{huang2026darl}, and the same method scores differently across papers that use different evaluation graders~\cite{yu2026rlpr,huang2026darl}, so the grader must be reported with the result.

For the rubric and judge rewards, verifier error and specification error must be separated, because a better judge fixes only the first. Section~\ref{sec:discussion} returns to this distinction.

\section{Discussion}\label{sec:discussion}
The four research questions point to one conclusion: the harm a verifier error does is set less by its rate than by whether the policy can find and reproduce it, and the evidence for this is strong in a narrow set of settings and thin outside them.

\subsection{Cross-Cutting Findings}
\textbf{F1. Structure decides the regime; rate decides the cost within it.} Exploitable false positives produce plateaus or collapse~\cite{egashira2026delay,zhao2025onetoken,li2026advjudge,tiwari2026prm,helff2026gaming}, random errors mostly delay learning~\cite{rad2026rate,plesner2026imperfect,lv2025climb}, and systematic errors the policy cannot trigger degrade accuracy gradually or stay bounded~\cite{zhu2026noisy,zhang2026leaky}. Earlier RL theory points the same way, since a few corrupted states can cause near-worst-case regret~\cite{everitt2017corrupted}. Preference-based RL shows a similar pattern: feature-dependent noise significantly degrades a state-of-the-art noise-robust method in some settings, where a method without explicit denoising often does better~\cite{li2026featurenoise}. The finding carries a qualifier, since under a fixed budget and a KL penalty, random errors also lower the endpoint, by an amount that depends strongly on the model~\cite{elmansouri2026noisegrpo}.

\textbf{F2. Error direction depends on what the checker checks and who built it.} Answer-matching rules mostly reject~\cite{huang2026accuracy,xu2026tinyv} and generative judges mostly accept~\cite{cai2026noisy,liu2025compassverifier}, but rules that test a proxy property accept~\cite{helff2026gaming}, tests written by a model mostly reject~\cite{rajan2026hackability}, and a reward model can be dominated by false negatives~\cite{elmansouri2026noisegrpo}. Because false positives cost more than false negatives at equal $J$~\cite{rad2026rate}, choosing rule rewards for their resistance to hacking~\cite{guo2025deepseekr1} is sound, but a move from rules to judges trades a benign error for a dangerous one only when the judge's false positives can be triggered.

\textbf{F3. Verifier error is non-stationary, and static accuracy is insufficient.} Learned and self-generated verifiers lose accuracy as the policy changes~\cite{cui2025prime,he2026unsupervised}, and shortcut use rises with reasoning effort and longer RL~\cite{helff2026gaming}. For reference-based checkers, this is a hypothesis, backed mainly by a reference-free verifier outside RL~\cite{zhou2026variation}. Static accuracy predicts training direction only coarsely~\cite{yan2026verifybench}: a PRM robust to gradient attacks was hacked in RL~\cite{tiwari2026prm}, a far more accurate verifier added only 0.8 points as reward~\cite{chen2025xverify}, and a less accurate checklist judge trained better policies than specialized reward models~\cite{viswanathan2025checklists}.

\textbf{F4. The evidence base is narrow and partly confounded.} Of the 46 studies that train a policy, 44 use a Qwen model, trained policies rarely exceed 32B parameters, and most tasks are mathematical. Several central results rest on a single seed or a single policy and judge pair~\cite{elmansouri2026noisegrpo,mahmoud2026rubrichacking,li2026advjudge,wang2026cherrl}, and contamination of Qwen2.5-Math on MATH-500 and AIME 2024~\cite{wu2026contamination} threatens every result obtained with that model on those benchmarks. Training studies of agentic verification do not yet exist; the evidence there comes from audits~\cite{rajan2026hackability,thaman2026rhb}.

\textbf{F5. Self-generated rewards are bounded by the prior and rescued by sparse verification.} Without an external check they sharpen and collapse at training-corpus scale~\cite{he2026unsupervised}, though they can help on small or test-time sets~\cite{zuo2025ttrl}, and their damage can hide in aggregate metrics~\cite{lin2026extinction,vakada2026labelfree}. A small amount of external verification recovers most of the gap~\cite{chen2026juryrl,wang2026selfbelief,liao2026t3rl}.

\subsection{Resolved and Open Disagreements}
Most apparent contradictions dissolve once the type of noise, the regime, or the budget is controlled. The claim that an imperfect verifier is good enough~\cite{plesner2026imperfect,lv2025climb} and the finding that noisy data is destructive~\cite{zhu2026noisy} concern different kinds of noise: random flips versus wrong labels that follow the model's own errors. The apparent collapse at 50\,\% flips~\cite{lv2025climb} is a failure to learn, as predicted at $J=0$~\cite{rad2026rate}. Large losses under random noise~\cite{elmansouri2026noisegrpo} need not contradict the mean-field result, whose recovery of the clean endpoint holds without regularization~\cite{rad2026rate}: that study used a fixed budget, a KL penalty and no ratio clipping, and under a KL penalty the dynamics settle at an interior equilibrium that depends on $J$~\cite{rad2026rate}, so longer training alone need not close the gap. 
Noise-aware corrections work on class-conditional noise~\cite{cai2026noisy,elmansouri2026noisegrpo} and fail on question-dependent labels~\cite{zhu2026noisy}, and within-group correlation of rule failures~\cite{xin2026independent} may reflect prompt difficulty rather than refute the independence those corrections assume.

Four disagreements remain open. The mechanism behind spurious-reward gains is contested~\cite{shao2026spurious,wu2026contamination,chen2026exploration}. Online self-certainty was stable over short runs~\cite{zhao2026intuitor} but collapsed over longer ones~\cite{zhang2026corewarding,he2026unsupervised}. Correlation-aware judge aggregation helped in one study~\cite{zhao2026care} and closed a little of the gap in another~\cite{kohli2026ninejudges}, and neither was tested in RL. Practitioners are advised both to prioritize precision~\cite{plesner2026imperfect} and to maximize $J$ rather than precision alone~\cite{rad2026rate}, while a verifier with 25 to 39\,\% precision still helped as a soft vote weight~\cite{liao2026t3rl}.

\subsection{A Recommended Reporting Protocol}
Results are hard to compare because studies rarely report the verifier in enough detail to reproduce its errors. Table~\ref{tab:protocol} lists recommended items, each motivated by a corpus result.

\begin{table}[t]
\caption{Recommended Reporting Protocol}
\label{tab:protocol}
\centering
\scriptsize
\setlength{\tabcolsep}{2.5pt}
\begin{tabularx}{\columnwidth}{@{}>{\raggedright\arraybackslash}p{1.6cm}Y>{\raggedright\arraybackslash}p{1.55cm}@{}}
\toprule
\textbf{Theme} & \textbf{Report} & \textbf{Motivation} \\
\midrule
Verifier & Identity, version, extraction configuration and coverage; judging granularity and output cap; the evaluation grader, separately & \cite{xin2026category,li2026verifybench,pan2026rubriceval,yu2026rlpr} \\
Verifier error & FPR and FNR, or $\hat J$ with an interval, on the policy's own outputs at the start and end of training; noise structure, with FP and FN tested separately; adversarial probes of model verifiers; invalid-response rate & \cite{rad2026rate,egashira2026delay,huang2026accuracy,liu2025compassverifier} \\
Training & An oracle reward beside the training reward; final and best checkpoints; pass@$k$ with pass@1; seeds or variance; no reliance on entropy as a health signal & \cite{egashira2026delay,zhu2026noisy,chen2026exploration} \\
Evaluation & More than one model family; random- and format-reward baselines; post-cutoff benchmarks and a contamination probe; per-problem and per-category breakdowns & \cite{shao2026spurious,wu2026contamination,lin2026extinction,vakada2026labelfree} \\
Label-free rewards & Reward accuracy against gold, not only pseudo-label accuracy & \cite{zuo2025ttrl,he2026unsupervised} \\
Data and release & Audited reference labels; distillation data filtered on true reward; released verdicts and code & \cite{zhu2026noisy,khalifa2026countdown,xin2026independent} \\
\bottomrule
\end{tabularx}
\end{table}

\subsection{Open Problems}
The largest gap is theoretical. The mean-field analysis of random errors~\cite{rad2026rate} has no counterpart for response-dependent, exploitable errors under group-relative gradients, although corrupted-reward theory offers a starting point~\cite{everitt2017corrupted}, and, as noted in Section~\ref{sec:background}, GRPO amplifies whatever the verifier accepts, false positives included. Empirically, verifier error should be measured on the policy's own outputs as training proceeds~\cite{zhou2026variation,cui2025prime}, the interaction between correlated errors and group size remains unexplored~\cite{xin2026independent,kohli2026ninejudges}, and defenses validated offline or on a single policy and judge pair need testing inside RL~\cite{zhao2025onetoken,zhao2026care,li2026advjudge}. Three further problems follow from the corpus: separating verifier error from rubric misspecification~\cite{mahmoud2026rubrichacking,gunjal2026rubrics}, guarding against hacking of the oracle itself~\cite{liu2026reasoningjudges}, and preventing hacking inherited through distillation data~\cite{khalifa2026countdown}. Agentic and long-horizon verification has been audited~\cite{rajan2026hackability,thaman2026rhb} but not yet studied in training.

\subsection{Limitations}
Our synthesis has two limitations. First, studies differ in models, budgets, and reference standards, so we compare directions and regimes, not effect sizes. Second, the search, based on keywords and four seed studies~\cite{egashira2026delay,plesner2026imperfect,xin2026category,xin2026independent}, covered English-language work up to September 2026 and may miss work framed in other terms.

\section{Conclusion and Future Work}\label{sec:conclusion}
This work surveyed 58 studies of verifier errors in RLVR. The evidence shows that the harm an error causes depends less on its rate than on whether the policy can find and reproduce it: exploitable false positives cause plateaus or collapse, random errors mostly slow learning, and errors that the policy cannot trigger degrade accuracy gradually or stay bounded. Whether a verifier errs by rejecting or accepting depends on what it checks and who built it, and its errors shift as the policy improves, so accuracy measured before training says little about robustness during it. Most mitigations have been tested against a single kind of error, and self-generated rewards remain bounded by the model's prior unless some external verification is added. We therefore argue for treating the verifier as a component to be measured on the policy's own outputs, probed adversarially, and reported in full. Open directions include a theory of exploitable errors, defenses validated inside RL across several policies and judges, and training studies of agentic verification.

\section*{Data Availability}
A list of all 81 cited works, with the 58 reviewed studies organized by research question and a link to each paper, is openly available under a CC~BY~4.0 license at \url{https://github.com/reasoning-ai-lab/Checking-the-Checkers-RLVR}.

\bibliographystyle{IEEEtran}
\bibliography{ref}

\end{document}